\documentclass[11pt]{article}

\usepackage[preprint]{acl}

\usepackage{times}
\usepackage{latexsym}

\usepackage[T1]{fontenc}

\usepackage[utf8]{inputenc}

\usepackage{microtype}

\usepackage{inconsolata}

\usepackage{graphicx}

\usepackage{amsmath}
\usepackage{pdfpages}
\usepackage{tabularx}
\usepackage{multirow}
\usepackage{booktabs}
\usepackage{subcaption}
\usepackage{amsfonts}

\title{Mind the Gap: Data Rewriting for Stable Off-Policy Supervised Fine-Tuning}

\author{
  Shiwan Zhao, Xuyang Zhao, Jiaming Zhou, Aobo Kong, Qicheng Li, Yong Qin\thanks{Corresponding Author.}\\
  College of Computer Science, Nankai University\\
  \texttt{zhaosw@gmail.com}\\  \texttt{\{xychao, zhoujiaming, kongaobo\}@mail.nankai.edu.cn}\\
  \texttt{\{liqicheng, qinyong\}@nankai.edu.cn} 
}

\begin{document}
\maketitle
\begin{abstract}
Supervised fine-tuning (SFT) of large language models can be viewed as an off-policy learning problem, where expert demonstrations come from a fixed behavior policy while training aims to optimize a target policy. Importance sampling is the standard tool for correcting this distribution mismatch, but large policy gaps lead to skewed weights, high variance, and unstable optimization. Existing methods mitigate this issue with KL penalties or clipping, which passively restrict updates rather than actively reducing the gap.
We propose a simple yet effective \emph{data rewriting} framework that proactively shrinks the policy gap before training. For each problem, correct model-generated solutions are kept as on-policy data, while incorrect ones are rewritten through guided re-solving, falling back to expert demonstrations only when needed. This aligns the training distribution with the target policy, reducing variance and improving stability. To handle residual mismatch after rewriting, we additionally apply importance sampling during training, forming a two-stage approach that combines \emph{data-level alignment} with lightweight \emph{optimization-level correction}.
Experiments on five mathematical reasoning benchmarks show consistent and significant gains over both vanilla SFT and the state-of-the-art Dynamic Fine-Tuning (DFT) approach. Data and code will be released at \url{https://github.com/NKU-HLT/Off-Policy-SFT}.
\end{abstract}

\section{Introduction}

Large language models (LLMs) have achieved remarkable progress in Chain-of-Thought (CoT) reasoning \cite{wei2022chain}, largely driven by a post-training pipeline that combines supervised fine-tuning (SFT) with reinforcement learning (RL) \cite{shao2024deepseekmath, lambert2024tulu, guo2025deepseek, liu2025acereason}.
SFT distills task-specific reasoning behaviors from high-quality demonstrations, enabling base models to rapidly adapt to novel tasks.  
RL complements this by optimizing on-policy rollouts with reward-based objectives, delivering consistent gains on challenging reasoning benchmarks. 
In the widely adopted \emph{SFT-then-RL} paradigm, SFT provides a strong initialization for reasoning that RL subsequently refines through on-policy sampling.

Despite their close connection, SFT and RL exhibit complementary strengths and limitations \cite{ma2025Interleave, yan2025offguidance}.  
SFT is simple and efficient, capable of expanding the model’s reasoning capability frontier by incorporating external expert knowledge and reasoning patterns.  
However, it operates entirely on off-policy data because expert demonstrations come from a fixed behavior policy rather than the evolving model policy, leading to the well-known policy gap and causing high variance, training instability, and overfitting.  
RL, in contrast, performs on-policy optimization and thus avoids the policy gap altogether, but it suffers from high sample and computational complexity and can only refine the model’s existing reasoning behaviors without introducing fundamentally new capabilities.  
In this work, we focus on improving SFT itself, providing a more stable foundation for standalone fine-tuning as well as for future extensions involving RL or hybrid approaches.

From the perspective of off-policy learning \cite{precup2000eligibility}, importance sampling (IS) is the standard tool for correcting the distribution mismatch between the behavior and target policies.  
However, when the policy gap becomes large, IS weights become highly skewed, leading to variance amplification and unstable optimization.  
Existing remedies, such as KL penalties, trust regions, or clipped ratios \cite{schulman2015trust, schulman2017proximal}, stabilize optimization by passively constraining updates but fail to actively reduce the underlying gap in the data distribution itself.

We propose a simple yet effective data rewriting framework that proactively reduces the policy gap before optimization begins. For each problem, we first sample multiple responses from the target model. If any response solves the problem correctly, we retain it as on-policy data. Otherwise, we prompt the model with the ground-truth solution as a reference to re-solve the problem, generating \emph{digest-and-retell} data that better reflects the target policy. If both self-solve and re-solve fail, we fall back to the original expert demonstration. Inspired by the intuition that true understanding emerges when learners re-express solutions in their own words rather than copying them verbatim, our \emph{digest-and-retell} strategy (see Figure \ref{fig:prompt}) transforms the SFT dataset to align it more closely with the target policy.

After rewriting, we apply a lightweight importance sampling correction during training to handle any residual distribution mismatch. This two-stage design—\emph{data-level alignment} through rewriting combined with \emph{optimization-level adjustment} via importance sampling—mitigates variance and stabilizes off-policy fine-tuning, ultimately leading to consistent performance gains across benchmarks.

Experiments on five mathematical reasoning benchmarks show that our approach consistently outperforms both vanilla SFT and the state-of-the-art Dynamic Fine-Tuning (DFT) approach \cite{wu2025DFT}.  
In particular, on the Qwen2.5-Math-7B model, our method improves the average accuracy from 23.23\% to 30.33\% over SFT and from 36.61\% to 42.03\% over DFT.

Our contributions are three-fold:
\begin{itemize}
    \item We formulate SFT as an off-policy learning problem and identify the policy gap as the key source of instability in IS-based optimization.
    \item We introduce a data rewriting framework that proactively reduces the policy gap at the data level, enabling low-variance and stable off-policy fine-tuning.
    \item We validate the approach across multiple models and benchmarks, demonstrating consistent gains over standard SFT and DFT baselines.
\end{itemize}

\section{Related Work}

\subsection{Data-Centric Improvements for SFT}
The quality of SFT largely depends on the construction of instruction datasets, and prior work explores three major aspects: scaling, diversity, and quality.  
Flan \cite{chung2024scaling, longpre2023flan} scales the number of instruction-tuning tasks and demonstrates that larger task collections substantially boost performance.  
LIMA \cite{zhou2023lima} shows that fine-tuning a strong pretrained model on only 1{,}000 carefully curated examples can achieve competitive results, highlighting the importance of data quality and diversity.  

Beyond scaling, several studies transform training examples to better align with the target model distribution.  
GRAPE \cite{zhang2025best} selects responses with the highest target-model likelihood from multiple LLMs before SFT training.  
Self-Distillation Fine-Tuning (SDFT) \cite{yang-etal-2024-self} generates distilled data using the model itself to bridge the distribution gap.  
Self-to-Supervised Fine-Tuning (S3FT) \cite{gupta-etal-2025-selective} identifies correct model responses and fine-tunes on them while paraphrasing or preserving gold answers for the remaining samples.  
Our data rewriting framework follows this line but adopts an off-policy perspective: it actively reduces the policy gap at the data level through rewriting and further mitigates residual mismatch via importance sampling during training.

\subsection{Combining SFT and RL}
Another line of work combines the capability expansion of SFT with the on-policy robustness of RL.  
Interleaved or unified training objectives jointly optimize supervised and reinforcement signals \cite{ma2025Interleave, liu2025uft, zhang2025Harmonizing}, while dynamic weighting strategies balance SFT imitation learning with RL-based preference optimization.  
Although RL operates on-policy, the SFT stage in these methods still relies on static off-policy datasets, leaving the underlying distribution mismatch unresolved.  
Our method is orthogonal: rather than modifying training objectives or interleaving on- and off-policy rollouts, it directly reduces the policy gap before optimization at the data level via data rewriting, offering a complementary perspective to existing objective-level approaches.

\subsection{Off-Policy Learning}
A parallel line of work views SFT as an off-policy learning problem, where expert demonstrations and the evolving target policy induce a distribution mismatch.  
One line of research introduces importance sampling or reward rectification to correct this mismatch \cite{wu2025DFT, qin2025SFTisRL}.  
Another line focuses on reducing the variance of importance sampling via optimization-level techniques such as clipping or trust regions \cite{zhu2025proximal}, which stabilize training by constraining or reweighting updates.  
In contrast, our data rewriting framework actively aligns the training distribution with the target policy before optimization, providing a complementary perspective to these optimization-level methods.

\section{Method}
\begin{figure}[t]
  \includegraphics[width=\columnwidth]{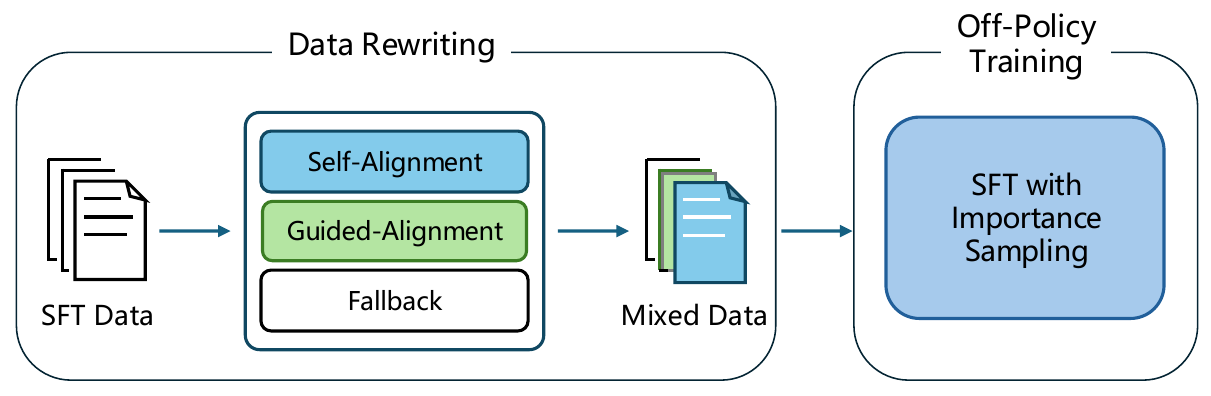}
\caption{The overall framework consists of (i) data rewriting, which converts SFT data from off-policy to a more on-policy distribution, and (ii) off-policy training with importance sampling, which further mitigates the remaining policy gap.}
\label{fig:framework}
\end{figure}

We view supervised fine-tuning (SFT) as an off-policy learning problem and propose a unified framework (Figure~\ref{fig:framework}) that combines \emph{data rewriting}, which proactively reduces the policy gap at the data level, with \emph{importance sampling} (IS), which further corrects residual mismatch during optimization.

\subsection{SFT as Off-Policy Learning}

Let $\pi_{sft}$ denote the behavior policy generating the SFT dataset, and let $\pi_\theta$ be the target policy parameterized by $\theta$. The goal of SFT is to maximize the expected reward under $\pi_\theta$:
\begin{equation}
    J(\theta) = \mathbb{E}_{(x,y)\sim \pi_\theta}[r(x,y)],
\end{equation}
where $r(x,y)$ is a task-specific reward signal (e.g., correctness).  

However, since training data come from $\pi_{sft}$ rather than $\pi_\theta$, SFT becomes an off-policy problem:
\begin{equation}
    J(\theta) = \mathbb{E}_{(x,y)\sim \pi_{sft}}\bigl[w(x,y)\, r(x,y)\bigr],
\end{equation}
where the importance weight is defined as
$w(x,y)=\frac{\pi_\theta(y\mid x)}{\pi_{sft}(y\mid x)}$. When the divergence $D(\pi_{sft}\,\|\,\pi_\theta)$ is large, these weights become highly skewed, leading to variance amplification and unstable optimization. Existing methods mitigate this issue using KL penalties, trust regions, or clipping, which passively constrain updates but do not actively reduce the underlying policy gap.

\subsection{Data Rewriting as Policy Alignment}
\begin{figure}[t]
  \includegraphics[width=\columnwidth]{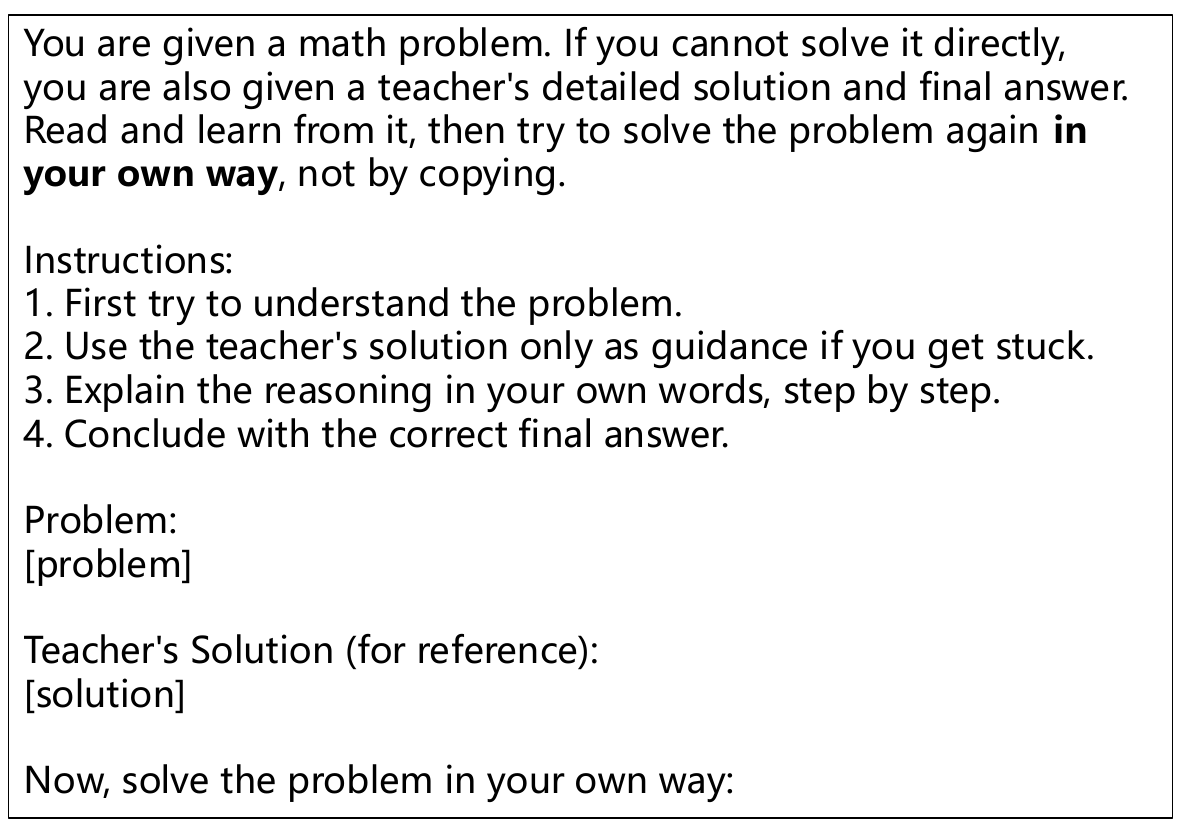}
  \caption{The \emph{digest-and-retell} prompt, which provides a reference solution and asks the model to re-solve the problem in its own words.}
  \label{fig:prompt}
\end{figure}

We introduce a data rewriting operator $\mathcal{T}$ that transforms $\pi_{sft}$ into a mixture distribution $\pi_{\text{mix}}$ before training:
\begin{equation}
    \pi_{sft} \xrightarrow{\;\mathcal{T}\;} \pi_{\text{mix}}
    \quad \text{with} \quad D(\pi_{\text{mix}}\|\pi_\theta) < D(\pi_{sft}\|\pi_\theta).
\end{equation}

The operator $\mathcal{T}$ applies a three-stage alignment hierarchy:
\begin{itemize}
    \item \textbf{Self-alignment:} For each input $x$, we sample multiple responses from $\pi_\theta$. If any response solves the problem correctly, we randomly retain one correct response as on-policy data\footnote{When multiple correct responses are available, one is randomly selected for fair comparison with standard SFT; the same strategy applies in the guided-alignment stage.}.
    \item \textbf{Guided-alignment:} For inputs where self-alignment fails, we prompt $\pi_\theta$ with reference solutions to generate \emph{digest-and-retell} responses that paraphrase the expert answers rather than copying them verbatim (see Figure~\ref{fig:prompt}). As in self-alignment, we sample multiple responses and, if any are correct, randomly retain one correct response as rewritten data.
    \item \textbf{Fallback:} If guided-alignment also fails, we fall back to the original expert demonstration.
\end{itemize}

The resulting dataset $\mathcal{D}^{\prime}$ consists of a mixture of on-policy and rewritten examples, with expert data included only as a fallback:
\begin{equation}
\mathcal{D}^{\prime} = \mathcal{D}_{\text{self}} \cup \mathcal{D}_{\text{retell}} \cup \mathcal{D}_{\text{expert}}.
\end{equation}

This hierarchical process ensures that $\pi_{\text{mix}} = \mathcal{T}(\pi_{sft})$ progressively shifts the training data toward the target policy, thereby reducing the policy gap before optimization begins.

\subsection{Importance Sampling with Aligned Data}

Although the resulting dataset $\mathcal{D}^{\prime}$ aligns more closely with the target policy distribution, residual mismatch may still remain due to imperfect rewriting and fallback expert demonstrations. Even with perfect rewriting, batch updates can introduce a policy gap as the target policy is progressively updated\footnote{We leave online data rewriting for each batch to future work.}. To mitigate this issue, we apply importance sampling during optimization:
\begin{equation}
\small
\mathcal{L}_{\mathrm{IS}}(\theta) 
= \mathbb{E}_{(x,y')\sim \mathcal{D}'}
\left[
-\sum_{t=1}^{|y'|}
w(x,y_t')\cdot
\log \pi_\theta(y_t' \mid x, y_{<t}')
\right],
\end{equation}
where  
\[
w(x,y_t')=\operatorname{sg}\!\left(
\frac{\pi_\theta(y_t' \mid x,y_{<t}')}
{\pi_{\mathrm{mix}}(y_t' \mid x,y_{<t}')}
\right)
\]
is the importance weight, and $\operatorname{sg}(\cdot)$ denotes the stop-gradient operator to prevent gradients from flowing through the weight itself.  
Following common practice \cite{wu2025DFT,zhang2025Harmonizing}, we approximate the denominator $\pi_{\mathrm{mix}}(y_t' \mid x,y_{<t}')\approx 1$, effectively treating the mixed data as the ground-truth distribution. 

This two-level design, combining \emph{data-level alignment} via rewriting with \emph{optimization-level correction} via IS, enables low-variance, unbiased training and serves as a complementary approach to methods such as Dynamic Fine-Tuning (DFT) \cite{wu2025DFT}.

\section{Experiments}

\begin{table*}[t]
\centering
\begin{tabular}{l c c c c}
\toprule
 & \textbf{Self-alignment} & \textbf{Guided-alignment} & \textbf{Fallback} & \textbf{Total} \\
\midrule
\textbf{Qwen2.5-Math-7B} & 28,752 & 11,620 & 7,634 & 48,006 \\
\textbf{Llama-3.1-8B-Instruct} & 26,947 & 16,335 & 4,719 & 48,001 \\
\bottomrule
\end{tabular}
\caption{Dataset statistics (number of instances) for different models across alignment stages.}
\label{tab:data}
\end{table*}

\subsection{Dataset and Models}
Following DFT \cite{wu2025DFT}, we use the NuminaMath CoT dataset \cite{numina_math_datasets} for training. The original dataset comprises approximately 860,000 mathematical problems along with their respective solutions. To reduce computational cost, we randomly sample 50,000 instances and retain around 48,000 after filtering out overlong examples.  

Since our method requires models to generate candidate solutions for data rewriting, we experiment with two representative backbones:  
\textbf{Qwen2.5-Math-7B} \cite{yang2024qwen2}, a math-specialized model without explicit instruction tuning, and  
\textbf{Llama-3.1-8B-Instruct } \cite{grattafiori2024llama}, a general instruction-tuned model.  
This comparison allows us to investigate whether our method can benefit both base and instruction-tuned models.

For data rewriting, we sample $10$ candidate responses from the target model in both the self-alignment and guided-alignment stages. Dataset statistics for both models are provided in Table~\ref{tab:data}. We observe that Qwen2.5-Math-7B solves more problems in the self-alignment stage but fewer in the guided-alignment stage, likely due to its weaker instruction-following capabilities compared to Llama-3.1-8B-Instruct.

\subsection{Training and Evaluation Details}

For SFT training, we use the \texttt{verl} framework \cite{sheng2025hybridflow} with the AdamW optimizer. The learning rate is set to $5\times 10^{-5}$ for Qwen2.5-Math-7B and $7\times 10^{-6}$ for Llama-3.1-8B-Instruct. Training is performed for one epoch with a batch size of 256. We employ a cosine decay learning rate schedule with a 0.1 warm-up ratio.

For evaluation, we follow DFT and assess mathematical reasoning performance on five widely used benchmarks: Math500 \cite{hendrycks2021measuring}, Minerva Math \cite{lewkowycz2022solving}, OlympiadBench \cite{ai-mathematical-olympiad-prize}, AIME 2024 \cite{aim-aime2024}, and AMC 2023 \cite{maa-amc2023}. All results are reported as the average accuracy over 16 decoding runs with a temperature of 1.0.

\subsection{Main Results}

\begin{table*}[t]
\centering
\begin{tabular}{l c c c c c c c}
\toprule
 & \textbf{Math500} & \textbf{Minerva} & \textbf{Olympiad} & \textbf{AIME24} & \textbf{AMC23} & \textbf{Avg.} \\

\toprule
    \textbf{Qwen2.5-Math-7B} & 39.90 & 14.43 & 17.16 & 7.50 & 29.38 & 21.67  \\
    + SFT & 52.61 & 19.13 & 17.32 & 2.06 & 25.00 & 23.23  \\
    + DFT & 68.70 & 31.92 & 32.31 & 6.68 & 43.44 & 36.61  \\
    + DR + SFT (ours) & 59.85 & 21.14 & 23.54 & 8.54 & 38.59 & 30.33  \\
    + DR + DFT (ours) & \textbf{70.40} & \textbf{34.85} & \textbf{36.12} & \textbf{14.58} & \textbf{54.22} & \textbf{42.03}  \\
\toprule
    \textbf{Llama-3.1-8B-Instruct} & 36.18 & 16.01 & 9.52 & 0.83 & 14.53 & 15.41  \\
    + SFT & 28.71 & 11.23 & 6.26 & 0.41 & 10.31 & 11.39  \\
    + DFT & 46.5 & 24.11 & 15.65 & 3.95 & 22.50 & 22.54  \\
    + DR + SFT (ours) & 44.38 & 19.21 & 13.07 & 1.87 & 17.19 & 19.14  \\
    + DR + DFT (ours) & \textbf{47.91} & \textbf{24.72} & \textbf{16.52} & \textbf{4.99} & \textbf{26.09} & \textbf{24.05}  \\
\bottomrule
\end{tabular}
\caption{Average accuracy (\%) on mathematical reasoning benchmarks. Our method, combined with either SFT or DFT, consistently improves performance over the corresponding baselines. DR stands for data rewriting.}
\label{tab:main_results}
\end{table*}

\begin{table*}[t]
\centering
\begin{tabular}{lccccc}
\toprule
\multirow{2}{*}{\textbf{Model}} & \multicolumn{2}{c}{\textbf{Self-alignment}} & \multicolumn{2}{c}{\textbf{Guided-alignment}} & \textbf{Fallback} \\
\cmidrule(lr){2-3} \cmidrule(lr){4-5} \cmidrule(lr){6-6}
 & SFT & Rewriting & SFT & Rewriting & SFT \\
\midrule
\textbf{Qwen2.5-Math-7B} & -184.44 & -83.36 & -280.64 & -259.44 & -296.03 \\
\textbf{Llama-3.1-8B-Instruct} & -193.01 & -127.67 & -341.57 & -303.44 & -388.17 \\
\bottomrule
\end{tabular}
\caption{Average log-probabilities of model responses across self-alignment, guided-alignment, and fallback subsets. Higher values (less negative) indicate responses closer to the target policy distribution.}
\label{tab:prob}
\end{table*}

\begin{table*}[t]
\centering
\begin{tabular}{lcccccc}
\toprule
Method & \textbf{Math500} & \textbf{Minerva} & \textbf{Olympiad} & \textbf{AIME24} & \textbf{AMC23} & \textbf{Avg.} \\
\midrule
DFT & 68.70 & 31.92 & 32.31 & 6.68 & 43.44 & 36.61 \\
DFT + Self-Alignment & 67.86 & 32.35 & 32.79 & 10.19 & 50.94 & 38.83 \\
DFT + Guided-Alignment & 70.06 & 32.57 & 32.16 & 8.54 &	48.44 &	38.36 \\
DFT + Full DR & \textbf{70.40} & \textbf{34.85} & \textbf{36.12} & \textbf{14.58} & \textbf{54.22} & \textbf{42.03} \\
\bottomrule
\end{tabular}
\caption{Ablation on Qwen2.5-Math-7B with DFT. \emph{Self-Alignment} replaces only correct model-generated responses, \emph{Guided-Alignment} replaces only digest-and-retell responses, and \emph{Full DR} combines both.}
\label{tab:ablation}
\end{table*}

Table~\ref{tab:main_results} reports results on five mathematical reasoning benchmarks. Our data rewriting (DR) method consistently improves both vanilla SFT and DFT on Qwen2.5-Math-7B and Llama-3.1-8B-Instruct.

\textbf{Qwen2.5-Math-7B.} On this math-specialized model, vanilla SFT achieves an average accuracy of 23.23\%, while DFT improves it to 36.61\%. Incorporating DR yields substantial gains, with DR+SFT increasing performance to 30.33\% and DR+DFT further boosting it to 42.03\% average accuracy. Notably, DR+DFT attains the highest scores across all five benchmarks, with particularly large improvements on AIME2024 (6.68\%~$\rightarrow$~14.58\%) and AMC2023 (43.44\%~$\rightarrow$~54.22\%).

\textbf{Llama-3.1-8B-Instruct.}  
On this instruction-tuned model, DR improves SFT from 11.39\% to 19.14\% and DFT from 22.54\% to 24.05\%. While gains remain consistent, the magnitude is smaller than that on Qwen2.5-Math-7B, suggesting that instruction-tuned models benefit less from data rewriting.

\begin{figure}[h!]  \includegraphics[width=\columnwidth]{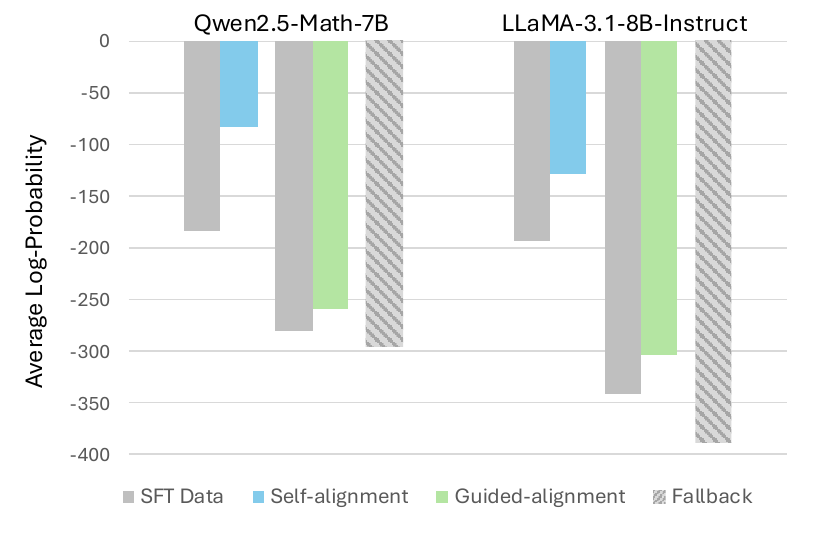}
\caption{Average log-probabilities for both models on Self-alignment, Guided-alignment, and Fallback subsets. SFT columns denote the original SFT data at each stage. In the Fallback stage, the Fallback subset coincides with the SFT data.}
\label{fig:prob}
\end{figure}

To better understand the overall performance gains and the performance gap between Qwen2.5-Math-7B and LLaMA-3.1-8B-Instruct, we analyze the average log-probabilities across the self-alignment, guided-alignment, and fallback subsets (Table~\ref{tab:prob}, Figure~\ref{fig:prob}). The analysis shows that rewriting effectively closes the policy gap, as rewritten responses consistently achieve higher average log-probabilities (less negative), indicating closer alignment with the target policy. Moreover, the SFT data across the three subsets reveal increasing problem difficulty, as reflected by decreasing log-probabilities from self-alignment to guided-alignment to fallback; the \emph{digest-and-retell} strategy still mitigates the policy gap on harder problems, although less effectively than self-solving. Finally, instruction tuning limits the benefits of data rewriting: LLaMA-3.1-8B-Instruct consistently exhibits lower log-probabilities than Qwen2.5-Math-7B across all subsets, suggesting that instruction tuning biases the model toward generic instruction-following and leaves less room for closing the policy gap.

\begin{figure*}[h!]
  \centering
  \begin{subfigure}{0.48\textwidth}
    \centering
    \includegraphics[width=\linewidth]{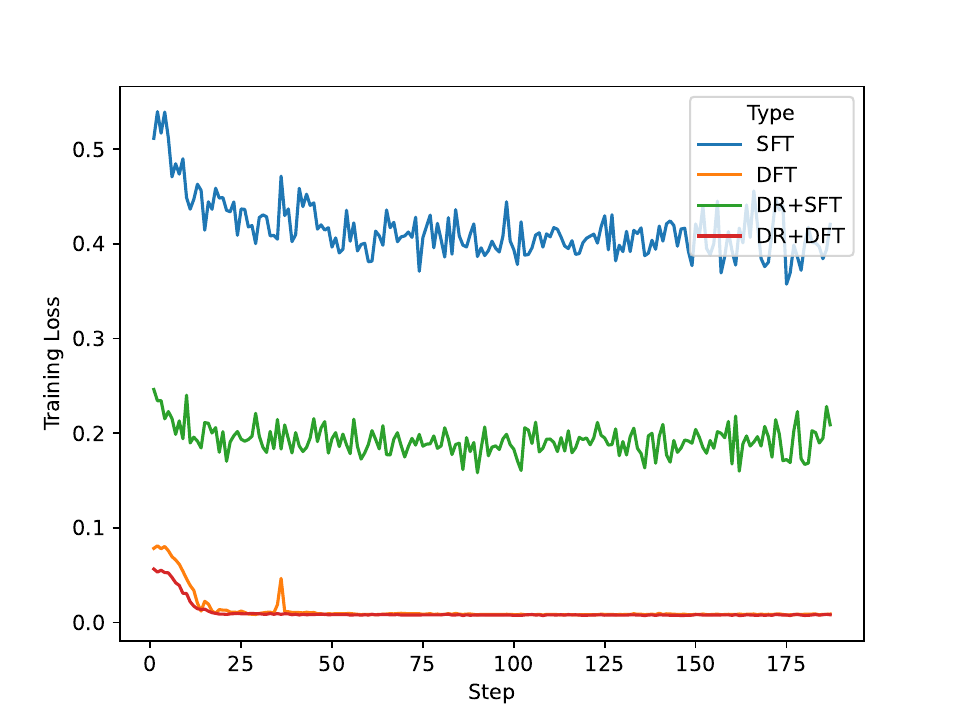}
    \caption{Qwen2.5-Math-7B}
  \end{subfigure}
  \hfill
  \begin{subfigure}{0.48\textwidth}
    \centering
    \includegraphics[width=\linewidth]{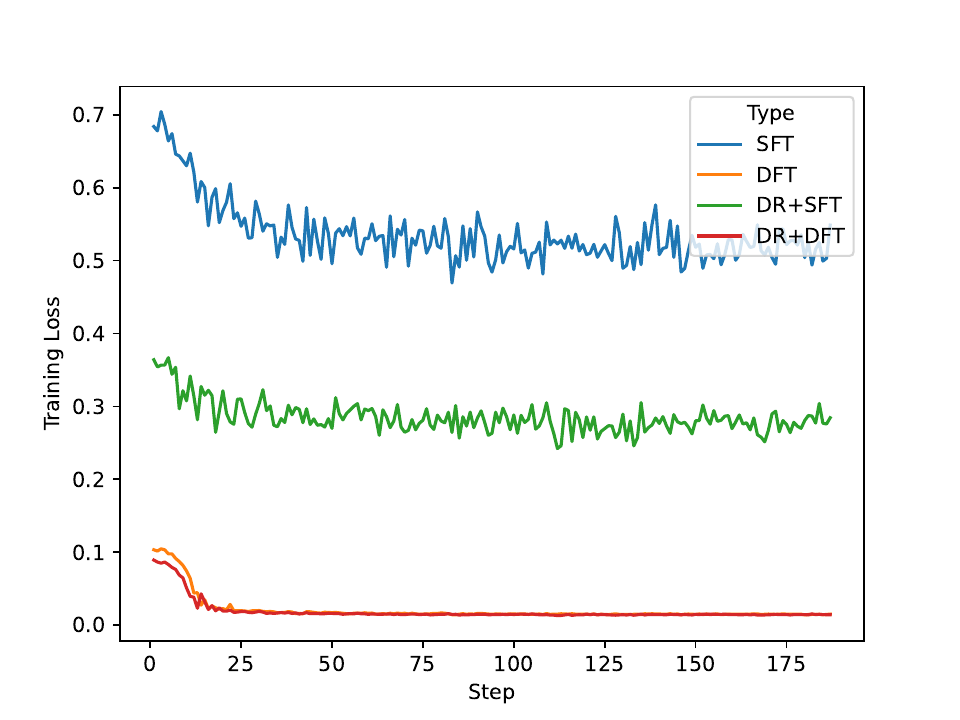}
    \caption{Llama-3.1-8B-Instruct}
  \end{subfigure}
  \caption{Training loss curves of Qwen2.5-Math-7B and Llama-3.1-8B-Instruct for SFT, DFT, and their combinations with data rewriting (DR). DR+DFT achieves the lowest final loss and the most stable convergence.}
  \label{fig:loss}
\end{figure*}

\noindent\textbf{Training Dynamics.}
Figure \ref{fig:loss} presents the training loss curves for both Qwen2.5-Math-7B and LLaMA-3.1-8B-Instruct across different methods. In both models, DFT and DR+DFT converge much faster than SFT and DR+SFT, reaching near-zero training loss within the first 40–50 steps, highlighting the strong supervision signal and optimization stability provided by dynamic fine-tuning. Data rewriting consistently reduces the loss of vanilla SFT on both models, confirming that aligning the training distribution with the target policy before optimization mitigates variance and stabilizes training. However, DR+SFT plateaus at a higher loss than DFT-based methods for both models, suggesting that residual distribution mismatch persists without dynamic on-policy updates. Notably, LLaMA-3.1-8B-Instruct exhibits consistently higher training loss than Qwen2.5-Math-7B across all methods, indicating a larger distribution mismatch between the training data and the instruction-tuned model, which explains its smaller performance gains from data rewriting. Finally, DR+DFT combines the benefits of both approaches, achieving the lowest final loss and the most stable convergence, accounting for its superior performance across all benchmarks in Table \ref{tab:main_results}.

\subsection{Ablation Study}

Table~\ref{tab:ablation} reports the ablation results on Qwen2.5-Math-7B with the DFT baseline. Incorporating \emph{Self-Alignment} into DFT yields moderate gains, raising average accuracy from 36.61\% to 38.83\% by replacing the original SFT data with correct model-generated responses. Adding \emph{Guided-Alignment}, which rewrites only the harder problems using the digest-and-retell strategy while keeping self-solved ones unchanged, also improves performance to 38.36\%, suggesting that guided rewriting effectively mitigates the policy gap when self-solving fails. Extending to \emph{Full DR}, which combines both self-alignment and guided-alignment, achieves the best results, boosting average accuracy to 42.03\% and delivering consistent gains across all benchmarks. These findings highlight the complementary roles of self-alignment and guided-alignment and the importance of proactively aligning training data before off-policy optimization to close the policy gap and enhance reasoning ability.

\section{Conclusion}
We presented a simple yet effective data rewriting framework for supervised fine-tuning of large language models, viewing SFT through the lens of off-policy learning. Our method proactively reduces the policy gap at the data level before training and further mitigates residual mismatch via importance sampling during optimization. Extensive experiments on multiple mathematical reasoning benchmarks show consistent and significant improvements over both vanilla SFT and the state-of-the-art Dynamic Fine-Tuning approach, with the largest gains achieved on base models. These results highlight the importance of data-centric approaches for stabilizing off-policy optimization and advancing the reasoning capabilities of large language models.

\section*{Limitations}

While our experiments demonstrate the effectiveness of data rewriting for stabilizing off-policy supervised fine-tuning, several limitations remain. First, our evaluation is restricted to a limited set of models, primarily at moderate parameter scales, so assessing its applicability to larger and more diverse models is left for future work. Second, we focus exclusively on mathematical reasoning benchmarks; extending the approach to broader domains, including industrial settings such as healthcare and finance, is an important next step. Third, our method adopts a single-round offline rewriting strategy, whereas more sophisticated or online approaches—e.g., rewriting per batch to mitigate policy shifts during training—could further enhance stability and performance. Finally, exploring richer rewriting techniques, such as leveraging external knowledge from more advanced models, represents another promising direction.

\bibliography{main}




\end{document}